\documentclass[runningheads,a4paper]{llncs}

\usepackage[utf8]{inputenc}
\usepackage{amsmath,amssymb,mathtools}
\usepackage[detect-weight=true, binary-units=true]{siunitx}
\usepackage{url,hyperref,cleveref}
\usepackage[linesnumbered]{algorithm2e}
\crefname{algocf}{alg.}{algs.}
\Crefname{algocf}{Algorithm}{Algorithms}
\usepackage[normalem]{ulem} 
\usepackage{color,colortbl,etoolbox,xcolor} 
\robustify\bfseries
\robustify\uline
\usepackage{multirow,booktabs,makecell,array}
\usepackage[caption=false]{subfig}
\usepackage{csquotes} 
\usepackage[inline]{enumitem}
\usepackage{hhline}

\usepackage{tikz,pgfplots,pgfplotstable}
\pgfplotsset{compat=newest}
\usetikzlibrary{decorations.pathmorphing,decorations.pathreplacing,arrows.meta,calc,intersections,through,hobby,positioning}
\usepgfplotslibrary{groupplots,statistics,fillbetween}

\title{Learning a Formula of Interpretability\\to Learn Interpretable Formulas}
\author{
    Marco Virgolin\inst{1} \and
    Andrea De Lorenzo\inst{2} \and
    Eric Medvet\inst{2} \and
    Francesca Randone\inst{3}
}
\institute{
    Centrum Wiskunde \& Informatica, Amsterdam, the Netherlands \and
    Department of Engineering and Architecture, University of Trieste, Trieste, Italy \and
    IMT School for Advanced Studies Lucca, Lucca, Italy
}

\begin{document}

\maketitle

\begin{abstract}
    Many risk-sensitive applications require Machine Learning (ML) models to be interpretable.
    Attempts to obtain interpretable models typically rely on tuning, by trial-and-error, hyper-parameters of model complexity that are only loosely related to interpretability.
    We show that it is instead possible to take a meta-learning approach: an ML model of non-trivial Proxies of Human Interpretability (PHIs) can be learned from human feedback, then this model can be incorporated within an ML training process to directly optimize for interpretability.
    We show this for evolutionary symbolic regression.
    We first design and distribute a survey finalized at finding a link between features of mathematical formulas and two established PHIs, \emph{simulatability} and \emph{decomposability}. 
    Next, we use the resulting dataset to learn an ML model of interpretability.
    Lastly, we query this model to estimate the interpretability of evolving solutions within bi-objective genetic programming.
    We perform experiments on five synthetic and eight real-world symbolic regression problems, comparing to the traditional use of solution size minimization. The results show that the use of our model leads to formulas that are, for a same level of accuracy-interpretability trade-off, either significantly more or equally accurate. Moreover, the formulas are also arguably more interpretable.
    Given the very positive results, we believe that our approach represents an important stepping stone for the design of next-generation interpretable (evolutionary) ML algorithms.
    
\end{abstract}

\section{Introduction}
Artificial Intelligence (AI), especially when intended as Machine Learning (ML), is increasingly pervading society.
Although ML brings numerous advantages, it is far from being fault-prone, hence its use comes with risks~\cite{adadi2018peeking,guidotti2018survey,lipton2016mythos,rudin2019stop}.
In many cases, failures with serious consequences could have been foreseen and prevented, if the ML models had not been unintelligible, i.e., \emph{black-boxes}.
Nowadays, especially for high-stakes applications, practitioners, researchers, and policy makers increasingly ask for ML to be used responsibly, fairly, and ethically~\cite{chouldechova2017fair,goodman2017european}.
Therefore, decisions taken by ML models need to be explainable~\cite{adadi2018peeking,arrieta2020explainable}. 

The field of eXplainable AI (XAI) studies techniques to provide explanations for the decisions taken by black-box models (or, more generally, AI systems), metrics that can be used as \emph{Proxies of Human Interpretability} (PHIs), as well as ML algorithms meant for the synthesis of models that are immediately interpretable~\cite{adadi2018peeking,rudin2019stop}.
In this paper, we consider the latter case. 

Several ML algorithms and techniques exist that are considered \emph{capable} of synthesizing interpretable models.
Among these, fitting linear models (e.g., by ordinary least squares or elastic net~\cite{zou2005regularization}), building decision trees~\cite{breiman1984classification}, and evolutionary program synthesis~\cite{poli2008field} are often listed in surveys on XAI (see, e.g., \cite{adadi2018peeking,guidotti2018survey}).
Unfortunately, in general, it cannot be \emph{guaranteed} that the model obtained by an ML algorithm will turn out to be interpretable.
For example, when a decision tree is built, the more the tree grows deep, the less the chances of the tree being interpretable.
Therefore, what is normally done is to repeat the ML training process (decision tree construction) with different hyper-parameter settings (tree depth) in a trial-and-error fashion, until a satisfactory model is obtained.
Trial-and-error, of course, comes with time costs. 
Next to this, another important issue is the fact that hyper-parameters are mostly meant to control the bias-variance interplay~\cite{bishop2006pattern}, and are but loosely related to interpretability. 

Multi-Objective Genetic Programming (MOGP) is a very interesting approach because, by its very nature, it mitigates the need for trial-and-error~\cite{poli2008field,zhao2007multi}. 
By evolving a population of solutions that encode ML models, MOGP can synthesize, in a single run, a plethora of models with trade-offs between accuracy and a chosen PHI.
Obtaining multiple models at once enhances the chance that a model with a satisfying trade-off between accuracy and interpretability will be found quickly.
Nonetheless, the problem of finding a suitable PHI remains.
So far, the PHI that have been used were quite simplistic.
For example, a common approach is to simply minimize the total number of model components (see \Cref{sec:related} for more). 
In this paper, we propose a way to improve upon the use of simplistic PHIs, and we focus on the case of MOGP for symbolic regression, i.e., where models are sought that can be written as mathematical formulas. 

Our proposal is composed of three main parts.
We begin by showing how it is possible to learn a model of non-trivial PHIs.
This can be seen as a concretization of an idea that was sketched in~\cite{doshi2017towards}: a data-driven approach can be taken to discover what features make ML models more or less interpretable. 
In detail,
\begin{enumerate*}[label=(\arabic*)]
    \item we design a survey about mathematical formulas, to gather human feedback on formula interpretability according to two established PHIs: \emph{simulatability} and \emph{decomposability}~\cite{lipton2016mythos} (see \Cref{sec:simu-deco});
    \item we process the survey answers and condense them into to a regression dataset that enables us to discover a non-trivial model of interpretability;
    \item we incorporate the so-found model within an MOGP algorithm, to act as an estimator for the second objective (the first being the mean squared error): in particular, the model takes as input the features of a formula, and outputs an estimate of interpretability.
\end{enumerate*}


All of our code, including the data obtained from the survey, is available at: \url{https://github.com/MaLeLabTs/GPFormulasInterpretability}.

\section{Related work}\label{sec:related}
In this paper, we focus on using ML to obtain interpretable ML models, particularly in the form of formulas and by means of (MO)GP.
We do not delve into XAI works where explanations are sought for the decisions made by a black-box model (see, e.g.,~\cite{ribeiro2016should,wang2017deep}), nor where the black-box model needs to be approximated by an interpretable surrogate (e.g., a recent GP-based work on this is~\cite{evans2019s}).
We refer to~\cite{adadi2018peeking,guidotti2018survey} as excellent surveys on various aspects of XAI. 
We describe the PHIs we adopt, and briefly mention works adopting them, in~\Cref{sec:simu-deco}.

Regarding GP for the synthesis of ML models, a large amount of literature is focused on controlling complexity, but not primarily as a means to enable interpretability.
Rather, the focus is on overfitting prevention, oftentimes (but not exclusively) by limiting bloat, i.e., the excessive growth of solution size~\cite{chen2018structural,poli2014parsimony,raymond2019genetic,sambo2020time,silva2012operator,vanneschi2010measuring,zhang1995balancing}.
Among these works, \cite{ekart2001selection,vladislavleva2008order,smits2005pareto} share with us the use MOGP, but are different in that they use hand-crafted complexity metrics instead of taking a data-driven approach (respectively solution size, order of non-linearity, and a modification of solution size), and again these metrics are designed to control bloat and overfitting instead of enable interpretability (\cite{vladislavleva2008order} does however discuss some effects on interpretability).

Among the works that use GP and focus on interpretability, \cite{cano2013interpretable} considers the evolution of rule-based classifiers, and evaluates them using a PHI that consists of dividing the number of conditions in the classifier by the number of classes. In~\cite{hein2018interpretable}, GP is used to evolve reinforcement learning policies as symbolic expressions, and complexity in interpretation is measured by accounting for variables, constants, and operations, with different ad-hoc weights.
The authors of~\cite{virgolin2019improving} study whether modern model-based GP can be useful when particularly compact symbolic regression solutions are sought, to allow interpretability.
A very different take to enable or improve interpretability is taken in~\cite{lensen2020genetic,tran2019genetic,virgolin2020explaining}, where interpretability is sought by means of feature construction and dimensionality reduction.
In~\cite{lensen2020genetic} in particular, MOGP is used, with solution size as a simple PHI.
Importantly, none of these works takes attempts to learn a PHI from data.

Perhaps the most similar work to ours is~\cite{mccormack2020understanding}.
Like we do, the authors train an ML model (a deep residual network~\cite{he2016deep}) from pre-collected human-feedback to drive an evolutionary process, but for a very different aim, i.e., automatic art synthesis (the human-feedback is aesthetic rankings for images).

\section{The survey}
We prepared an online survey (\url{http://mathquiz.inginf.units.it}) to assess the simulatability and decomposability of mathematical formulas (we referred to~\cite{burgess2001guide} for survey-preparation guidelines).
We begin by describing the two PHIs, and proceed with an overview of the content of the survey and the generation process. We provide full details on online supplementary material at: \url{https://github.com/MaLeLabTs/GPFormulasInterpretability}.


\subsection{Simulatability and decomposability}\label{sec:simu-deco}
Simulatability and decomposability are two established PHIs, introduced in a seminal work on XAI~\cite{lipton2016mythos}.
Simulatability represents a measurable proxy for the capability of a person to contemplate an entire ML model, and is measured by assessing whether a human, given some input data, can reproduce the model's output within a reasonable error margin and time~\cite{lipton2016mythos}.
No specific protocol exists to perform the measurement.
In~\cite{poursabzi2018manipulating}, this PHI was measured as the absolute deviation between the human estimate for the output of a (linear) model and the actual output, given a set of inputs.
With our survey, we measured the rate of correct answers to multiple choices questions on the output of a formula.

Decomposability represents the possibility that a model can be interpreted by parts: inputs, parameters, and (partial) calculations of a model need to admit an intuitive explanation~\cite{lipton2016mythos}.
For example, the coefficients of a linear model can be interpreted as the strengths of association between features and output.
Decomposability is similar to the concept of \emph{intelligibility} of~\cite{lou2012intelligible}. 
As for simulatability, there exists no prescription on how to measure decomposability.
We considered variables as important components to represent this PHI, and gathered answers (correct/wrong) on properties of the behavior of a formula when one of its variables varies within an interval.

\subsection{Overview on the survey and results}
\label{sec:survey}
We implemented the survey as a webpage, consisting of an introductory section to assess the respondents' level of familiarity with formulas, followed by eight questions, four about simulatability, and four about decomposability.
The eight questions are randomly selected when the webpage is loaded, from a pre-generated database that contains \num{1000} simulatability and \num{1000} decomposability questions, each linked to one of \num{1000} automatically generated formulas. 
\Cref{fig:example-questions} shows examples of these questions.
Each and every question presents four possible answers, out of which only one is correct.
Alongside each question, the user is asked to state how confident (s)he is about the answer, on a scale from \num{1} to \num{4}.

\begin{figure}[t]
    \centering
    \begin{tabular}{|l|c|l|}
    \hhline{|-|~|-|}
    \makecell[l]{ 
    \scriptsize Given the formula \color{blue} $\frac{5x_1 + 1}{\cos(x_2 - 3.14)}$ \color{black} and the\\
    \scriptsize input value(s) \color{blue} $[ x_1 = 8.0, x_2 = 6.28 ]$\color{black}, which \\
    \scriptsize option is closest to the output? \\
    \scriptsize (a) \num{-410.0} \\ 
    \scriptsize (b) \num{-41.0} \\ 
    \scriptsize (c) \num{410.0} \\ 
    \scriptsize (d) \num{-20.5} \\ 
    }
    &&
    \makecell[l]{ 
    \scriptsize Consider the formula \color{blue} $5\sin^{0.5}(x-3.14)$ \color{black}. \\ 
    \scriptsize Which option best describes the behavior\\
    \scriptsize of the function as \color{blue} $x$ \color{black} varies in \color{blue} $[-1.0,1.0]$\color{black}? \\
    \scriptsize (a) The function is bounded but not always defined \\ 
    \scriptsize (b) The function is not bounded nor always defined \\ 
    \scriptsize (c) The function is not bounded but always defined \\ 
    \scriptsize (d) The function is bounded and always defined \\ 
    }
    \\
    \hhline{|-|~|-|}
    
    \end{tabular}
    
    \caption{
        Examples of questions on simulatability (left) and decomposability (right).
    }
    \label{fig:example-questions}
\end{figure}

The \num{1000} formulas were encoded with trees, and randomly generated with a half-and-half initialization of GP~\cite{poli2008field} (max depth \num{4}). The set of leaf nodes for the trees included \num{4} different variables, and constants that were either integers (from 0 to 10) or multiples of $\pi$ ($3.14$ or $6.28$).
The possible operations were $+$, $-$, $\times$, $\div$, $^\wedge$, $\sqrt{\cdot}$, $\sin$, and $\cos$.
We performed rejection sampling and automatic simplifications to avoid presenting fundamentally uninteresting functions (e.g., constant ones), or functions with exploding output (e.g., due to $^\wedge$). 

For simulatability questions, the user was either asked to pick the correct 2D graph representing the behavior of the (one variable) formula, or to choose the best estimate of the output of the (multi-variable) formula, given values for the variables.
Decomposability questions asked whether the formula was (not) bounded, (not) always defined, (not) null in some points, (not) negative in some points, for one variable changing in a given interval and the others being fixed.

\begin{figure}[t]
    \centering
    \begin{tikzpicture}
        \begin{groupplot}[
            width=0.4\linewidth,
            height=0.19\linewidth,
            xbar,
            enlarge y limits=0.2,
            enlarge x limits=0.02,
            title style={align=left,yshift=-3mm,font=\tiny},
            ticklabel style={font=\tiny},
            /pgf/bar width=1.25mm,
            group style={
                group size=2 by 2,
                horizontal sep=22mm,
                vertical sep=5mm,
                xticklabels at=edge bottom
            }
        ]
            \nextgroupplot[
                title={How frequently do you deal with mathematical\\expressions in your work or study?},
                symbolic y coords = {$<$weekly,weekly,daily},
            ]
            \addplot coordinates { (0.07784431,$<$weekly) (0.15568862,weekly) (0.76646707,daily) };
            \nextgroupplot[
                title={How long have you been working and/or\\studying ina math related field?}, 
                symbolic y coords = {$<$month,$<$year,$1$--$3$ years,$>$3 years}
            ]
            \addplot coordinates { (0.06586826,$<$month) (0.04790419,$<$year) (0.16467066,$1$--$3$ years) (0.72155689,$>$3 years) };
            \nextgroupplot[
                title={What is the complexity of\\the formulas you usually deal with?},
                symbolic y coords = {fairly complex,moderately complex,simple}
            ]
            \addplot coordinates { (0.17664671,simple) (0.51796407,moderately complex) (0.30538922,fairly complex)};
            \nextgroupplot[
                title={How well do you deal with complex formulas?},
                symbolic y coords = {bad w/ any,bad w/ complex,good w/ any}
            ]
            \addplot coordinates {(0.05988024,bad w/ any) (0.46407186,bad w/ complex) (0.4760479,good w/ any)};
        \end{groupplot}
    \end{tikzpicture}    
    \vspace{-.5cm}
    \caption{
        Distribution of answers about user expertise.
    }
    \label{fig:results-survey-results}
\end{figure}
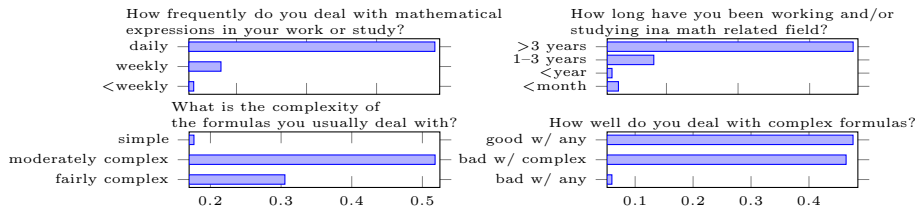

We distributed the survey by emailing research groups and departments within the institutes of the authors, targeting both students and faculty members.
We further shared the survey on Reddit (subreddit CasualMath) and Twitter.  
We obtained \num{334} responses in $\approx 35$ days, corresponding to \num{2672} answers.
\Cref{fig:results-survey-results} shows the distribution of answers to the introductory part of the survey.

\section{Learning a formula of interpretability}
We now describe how we condense the survey answers into a regression dataset, and use this dataset to learn an ML model (as a formula) of interpretability.

We begin with replacing each question with a set of feature values that represents the formula contained in the question (explained in detail below).
We obtain multiple identical sets of feature values with different outcomes in terms of correctness and confidence.
We merge equal sets of feature values into a single sample, taking the ratio of correct answers and the mean confidence.
In doing so, we do not distinguish between answers belonging to simulatability or decomposability, assuming they are equally good PHIs.
We also remark that we did not make expertise-based partitions because of the limited number of respondents and the skew in expertise distribution (\Cref{fig:results-survey-results}).

As label to regress, we take the product between correctness ratio and confidence (the latter normalized to have values $\frac{0}{3}$, $\frac{1}{3}$, $\frac{2}{3}$, $\frac{3}{3}$).
We choose to weight by confidence because, arguably, the less a user is confident about the answer, the less (s)he feels (s)he interprets the formula correctly.
Essentially, this is a new PHI synthesized out of simulatability and decomposability, that takes confidence into account.
From now on, we refer to this PHI as $\phi$.

The choice of what formula features are considered is of crucial importance as it determines the way the answers are merged. 
We ultimately consider the following features:
the size $\ell$ of the formula (counting variables, constants, and operations),
the number $n_\text{o}$ of operations,
the number $n_\text{nao}$ of non-arithmetic operations,
the number $n_\text{naoc}$ of consecutive compositions of non-arithmetic operations.
Note that the number of variables or constants is $\ell - n_\text{o}$, and the number of arithmetic operations is $n_\text{o} - n_\text{nao}$. 

By merging answers sharing the same values for the aforementioned four features, and excluding merged samples composed by less than \num{10} answers for robustness, we obtain a small regression dataset with \num{73} samples.

\subsection{Learning the model}
\begin{figure}[b]
    \centering
    \subfloat[\label{fig:phi-dist}Histogram of $\phi$ values.]{
        \begin{tikzpicture}
            \begin{axis}[
                ybar,
                ymin=0,
                width=0.5\linewidth,
                height=0.35\linewidth,
                ylabel={\% of samples},
                xtick={0.3,0.4,0.5,0.6,0.7,0.8,0.9},
                xtick align=inside
            ]
                \addplot table [x=phi, y expr=\thisrowno{1}*100] {phi-summary.txt};
            \end{axis}
        \end{tikzpicture}
    }
    \subfloat[\label{fig:coeffs-boxplot}Boxplots of learned coefficients.]{
        \begin{tikzpicture}
            \begin{axis}[
                width=0.5\linewidth,
                height=0.35\linewidth,
                ytick={1,2,3,4,5,6},
                yticklabels={$\ell$ , $n_\text{o}$ , $n_\text{nao}$ , $n_\text{naoc}$}
            ]
                \addplot [boxplot] table [y=n_nodes] {coeff.txt};
                \addplot [boxplot] table [y=n_ops] {coeff.txt};
                \addplot [boxplot] table [y=na_ops] {coeff.txt};
                \addplot [boxplot] table [y=na_comp] {coeff.txt};
            \end{axis}
        \end{tikzpicture}
    }
    \caption{
        Salient information about the learning data and the linear model.
    }
    \label{fig:model}
\end{figure}
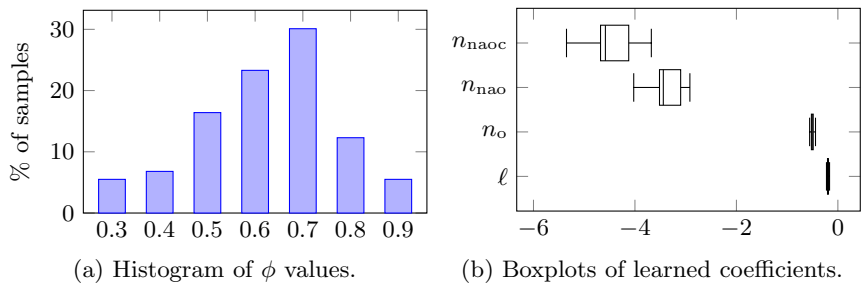

\Cref{fig:phi-dist} shows the distribution of $\phi$. 
Since this distribution is not uniform, similarly to what is done for classification with imbalanced class frequency, we weight samples by the inverse frequency of the bin they belong to.

To obtain a readable ML model and due to the small number of samples, we choose to fit a elastic net linear model~\cite{zou2005regularization} of the four features with stochastic gradient descent, and validate it with leave-one-out cross-validation.
We refer the reader interested in the details of this process (which includes, e.g., hyper-parameter tuning) to \url{https://github.com/MaLeLabTs/GPFormulasInterpretability}.
The leave-one-out cross-validation scores a (weighted) training $R^2=0.506$, and (weighted) test $R^2=0.545$ (mean weighted absolute error of \SI{26}{\percent}).
The distribution of the model coefficients optimized across the folds is shown in \Cref{fig:coeffs-boxplot}.

We take the average coefficients found during the cross-validation to obtain the final model of $\phi$ (from now on, considered as a percentage):
\begin{align}
    \label{eq:model-phi}
    \mathcal{M}^\phi (\ell , n_\text{o}, n_\text{nao} , n_\text{naoc}) = & 79.1 -0.2 \ell -0.5 n_\text{o} -3.4 n_\text{nao} -4.5 n_\text{naoc}.
\end{align}
By observing $\mathcal{M}^\phi$, it can be seen that each feature plays a role in lowering interpretability, yet by different magnitudes; $n_\text{naoc}$ is the most important factor.

\section{Exploiting the model of interpretability in MOGP}
The experimental setup adopted for the use of the model $\mathcal{M}^\phi$ within an MOGP algorithm for symbolic regression is presented next.
We describe the algorithm we use, its objectives, the datasets we consider, and the evaluation process.

\subsubsection{MOGP by NSGA-II.}
We use a GP version of NSGA-II~\cite{deb2002fast}, the most popular multi-objective evolutionary algorithm, and refer to it as \emph{NSGP} (the same name has been used in different works, e.g.,~\cite{liang2016multi,wang2014multiobjective,watchareeruetai2009construction}). 
We use traditional settings (all described in~\cite{poli2008field}): tree-based encoding; ramped half-and-half initialization (min and max depth of \num{1} and \num{6} respectively); and tournament selection (size \num{2}, default in NSGA-II). 
The crossover operator is subtree crossover (probability of \num{0.9}, default in NSGA-II).
The mutation operator is one-point mutation (probability of $1/\ell$ for each tree node, with $\ell$ the number of nodes). 

We set the population size to \num{1000} and perform \num{100} generations across all experiments.
The possible tree leaves are the problem variables and an ephemeral random constant~\cite{poli2008field}, with random values from $\mathcal{U}(-5, +5)$. 
The operations are $+$, $-$, $\times$, $\div_p$, $\sin$, $\cos$, $\exp$, and $\log_p$. 
Protection of division by zero is implemented by $\div_p( i_1, i_2 ) := \text{sign}(i_2) \frac{ i_1 }{ |i_2| + \epsilon }$. 
Similarly, the logarithm is protected by taking as argument the absolute value of the input plus $\epsilon$. 
We use $\epsilon=10^{-6}$. 
Trees are not allowed to grow past $100$ nodes, as they would definitely be not interpretable.

Our (Python 3) implementation of NSGP (including an interface to scikit-learn~\cite{scikit-learn}) is available at: \url{https://github.com/marcovirgolin/pyNSGP}.

\subsubsection{Objectives.}
We consider two competing objectives: error vs.\ interpretability.
For the first objective, we consider the Mean Squared Error (MSE) with \emph{linear scaling}~\cite{keijzer2003improving,keijzer2004scaled}, i.e., $\text{MSE}^\text{lin.scal.}(y,\hat{y}) = \frac{1}{N} \sum_{i=1}^{N} \left(y_i - a - b \hat{y_i}\right)^2$.
The use of the optimal (on training data) affine transformation coefficients $a,b$ corresponds to computing an absolute correlation.
From now on, we simply refer to this as MSE.

For the second objective, we consider two possibilities. 
The first one is realized by using our model $\mathcal{M}^\phi$: we extract the features from the tree to be evaluated, feed them to $\mathcal{M}^\phi$, and take the resulting estimate of $\phi$. 
To conform with objective minimization, we actually seek to minimize the opposite of this estimate (we also ignore the intercept term of~\Cref{eq:model-phi}). 
We call $\text{NSGP}^\phi$ the version of NSGP using this second objective.

The second possibility is to solely minimize the number of nodes $\ell$. 
This approach, despite its simplicity, is very popular (see \Cref{sec:related}).
Here we use it as a baseline for comparison. 
We refer to NSGP using $\ell$ minimization as $\text{NSGP}^\ell$. 

The objectives based on $\ell$ and $\phi$ are in a comparable scale (considering $\phi$ as a percentage).
To bring the first objetive to a similar scale (since scale impacts the crowding distance~\cite{deb2002fast}), during the evolution we multiply the MSE by $\frac{100}{\sigma^2(y)}$ (i.e., predicting the mean $\mu(y)$ achieves an error of \num{100}).

NSGP measures the quality of solutions according to the same criteria of~\cite{deb2002fast} (domination ranking  and crowding distance). 
We will report results relative to the front of non-dominated solutions $\mathcal{F}$ obtained at the end of the run. 
Recall that a solution is non-dominated if there exists no other solution that is better in at least one objective and not worse in all others.
In other words, $\mathcal{F}$ contains the solutions with best trade-offs between the objectives.

\subsubsection{Datasets and evaluation.} 
We consider \num{13} regression datasets in total, see Table~\ref{tab:datasets}. 
The first \num{5} datasets are synthetic (S-) and are recommended in~\cite{white2013better}. 
The other \num{8} regard real-world data (R-) and are (mostly) taken from the UCI machine learning repository~\cite{dua2017uci} and used in recent literature (e.g.,~\cite{ruberto2020sgp,virgolin2019linear}).

We treat all datasets the same. 
We apply standardization, i.e., all features are set to have zero mean and unit standard deviation. 
Before each run, we partition the dataset in exam at random, splitting it into \SI{80}{\percent} samples for training, \SI{10}{\percent} for validation, and \SI{10}{\percent} for testing. 
The training set is used by NSGP to evolve the solutions. 
The other sets are used to assess generalization, as is good practice in ML~\cite{bishop2006pattern}. 
In particular, using the final population, we re-compute the MSE of the solutions w.r.t.\ the validation set, and compute the front of non-dominated solutions $\mathcal{F}$ based on this.
The MSE of the solutions in this front is finally re-evaluated on the test set (test MSE).

Because dataset partitioning as well as NSGP are stochastic, we perform \num{50} runs per dataset. 
To evaluate whether differences in results between $\text{NSGP}^\phi$ and $\text{NSGP}^\ell$ are significant, we use the Wilcoxon signed-rank test~\cite{demvsar2006statistical} to the \SI{95}{\percent} confidence level, including Holm-Bonferroni correction~\cite{holm1979simple}.

\begin{table}[]
    \caption{
        Datasets used in this work.
        For the synthetic datasets, $N$ is chosen by considering the largest between the training set and test set proposed in~\cite{white2013better}.
    }
    \centering
    \begin{minipage}[t]{.485\linewidth}
        \centering
        \begin{tabular}[t]{ll S[table-format=5.0] S[table-format=2.0] S[table-format=1.2] S[table-format=1.2]}
            \toprule
            Dataset &  Abbr. & {$N$} & {$D$} & {$\mu(y)$} & {$\sigma(y)$} \\
            \midrule
            Keijzer 6 & S-Ke6 & 121 & 1 & 4.38 & 0.98 \\
            Korns 12 & S-K12 & 10000 & 5 & 2.00 & 1.06\\
            Nguyen 7 & S-Ng7 & 20 & 1 & 0.79 & 0.48\\
            Pagie 1 & S-Pa1 & 625 & 2 & 1.56 & 0.49\\
            Vladislav.\ 4 & S-Vl4 & 5000 & 5 & 0.49 & 0.19\\
            \bottomrule
        \end{tabular}
    \end{minipage}\hfill%
    \begin{minipage}[t]{.505\linewidth}
        \centering
        \begin{tabular}[t]{ll S[table-format=4.0] S[table-format=2.0] S[table-format=3.1] S[table-format=2.1]}
            \toprule
            Dataset &  Abbr. & {$N$} & {$D$} & {$\mu(y)$} & {$\sigma(y)$} \\
            \midrule
            Airfoil & R-Air & 1503 & 5 & 124.8 & 6.9\\
            Boston housing & R-Bos & 506 & 13 & 22.5 & 9.2\\ 
            Dow chemical & R-Dow & 1066 & 57 & 3.0 & 0.4\\
            Diabetes & R-Dia & 442 & 10 & 152.1 & 77.0\\
            Energy cooling & R-EnC & 768 & 8 & 24.6 & 9.5\\
            Energy heating & R-EnH & 768 & 8 & 22.3 & 10.1\\
            Tower & R-Tow & 4999 & 25 & 342.1 & 87.8\\
            Yacht & R-Yac & 308 & 6 & 10.5 & 15.1\\
            \bottomrule
        \end{tabular}
    \end{minipage}
    \label{tab:datasets}
\end{table}
\vspace{-1cm}

\section{Results}

\subsubsection{Fitting and generalization error.}
We begin by reporting quantitative results of the models in terms of training and test MSE.
Although the test MSE is what ultimately matters in practical applications (i.e., a good formula is one that generalizes to unseen data), we also show the training MSE because it reflects the capability of an algorithm to optimize as much as possible. 
We present results for different trade-off levels $\tau$. 
Specifically, $\tau$ is the percentile rank of the solutions in the non-dominated front $\mathcal{F}$ ordered by increasing MSE: $\tau=1$ considers the solution with best MSE and worst PHI; $\tau=100$ considers the solution with worst MSE and best PHI (see \Cref{fig:scatters}). 
\Cref{tab:percentiles-scores} shows the MSE obtained by $\text{NSGP}^\phi$ and $\text{NSGP}^\ell$ at training and test times, alongside the values of $\phi$ and $\ell$, for the MSE-specialized part of the fronts ($\tau=5,25,50$).

\begin{table}[]
    \caption{
        Median performance from 50 runs of the solutions found by $\text{NSGP}^\phi$ and $\text{NSGP}^\ell$ at different trade-off levels $\tau$ ($\tau=1$ for best MSE, $\tau=100$ for best PHI).
        Median front sizes ($|\mathcal{F}|$)  are computed w.r.t.\ the validation set. 
        MSE values in bold for one version of NSGP are significantly lower than the corresponding ones for the other version of NSGP at the \SI{95}{\percent} confidence level after Holm-Bonferroni correction. \label{tab:percentiles-scores}
    }
    \centering
    \begin{tabular}{
        l@{\hspace{2.5mm}}
        S[table-format=2.0]@{\hspace{2.5mm}}
        S[table-format=1.3]S[table-format=1.3]S[table-format=2.1]S[table-format=2.0]c@{\hspace{2.5mm}}
        S[table-format=1.3]S[table-format=1.3]S[table-format=2.1]S[table-format=2.0]c@{\hspace{2.5mm}} 
        S[table-format=1.3]S[table-format=1.3]
    }
        \toprule
        &
        & \multicolumn{5}{c}{$\text{NSGP}^\phi$}
        & \multicolumn{5}{c}{$\text{NSGP}^\ell$} \\
        \cmidrule(lr){3-7}
        \cmidrule(l){8-12}
        && {Train} & {Test} &&& 
        & {Train} & {Test} &&& 
        & {Train} & {Test} \\
        Dataset & {$\tau$} &
        {MSE} & {MSE} & {$\phi$} & {$\ell$} & {$|\mathcal{F}|$} &
        {MSE} & {MSE} & {$\phi$} & {$\ell$} & {$|\mathcal{F}|$} &
        {$p$-value} & {$p$-value}\\
        \midrule
\multirow{3}{*}{S-Ke6} & 5 & \bfseries 0.000 & \bfseries 0.001 & 11.4 & 11 & \multirow{3}{*}{7} & 0.007 & 0.006 & 14.5 & 8 & \multirow{3}{*}{5} & 0.000 & 0.000 \\
 & 25 & \bfseries 0.001 & \bfseries 0.002 & 9.4 & 8 &  & 0.013 & 0.007 & 13.5 & 6 &  & 0.000 & 0.000 \\
 & 50 & \bfseries 0.005 & \bfseries 0.007 & 3.8 & 7 &  & 0.023 & 0.023 & 7.4 & 4 &  & 0.000 & 0.000 \\
\midrule
\multirow{3}{*}{S-K12} & 5 & \bfseries 0.997 & 0.998 & 2.9 & 7 & \multirow{3}{*}{3} & 0.998 & 0.997 & 7.4 & 4 & \multirow{3}{*}{2} & 0.000 & 0.924 \\
 & 25 & \bfseries 0.998 & 0.998 & 2.9 & 7 &  & 0.998 & 0.997 & 7.4 & 4 &  & 0.000 & 0.941 \\
 & 50 & \bfseries 0.998 & 0.997 & 2.0 & 5 &  & 0.998 & 0.997 & 7.4 & 3 &  & 0.000 & 0.454 \\
\midrule
\multirow{3}{*}{S-Ng7} & 5 & \bfseries 0.000 & \bfseries 0.000 & 4.7 & 9 & \multirow{3}{*}{4} & 0.004 & 0.003 & 12.6 & 4 & \multirow{3}{*}{2} & 0.000 & 0.000 \\
 & 25 & \bfseries 0.001 & \bfseries 0.001 & 2.9 & 7 &  & 0.005 & 0.003 & 12.6 & 4 &  & 0.000 & 0.000 \\
 & 50 & \bfseries 0.001 & \bfseries 0.001 & 2.0 & 5 &  & 0.005 & 0.003 & 12.6 & 3 &  & 0.000 & 0.000 \\
\midrule
\multirow{3}{*}{S-Pa1} & 5 & \bfseries 0.174 & \bfseries 0.190 & 15.9 & 16 & \multirow{3}{*}{10} & 0.216 & 0.221 & 22.8 & 7 & \multirow{3}{*}{6} & 0.000 & 0.001 \\
 & 25 & 0.221 & 0.231 & 14.1 & 12 &  & 0.257 & 0.269 & 19.7 & 6 &  & 0.038 & 0.004 \\
 & 50 & 0.396 & 0.392 & 10.5 & 8 &  & 0.338 & 0.387 & 13.5 & 5 &  & 0.029 & 0.950 \\
\midrule
\multirow{3}{*}{S-Vl4} & 5 & 0.509 & 0.536 & 13.9 & 9 & \multirow{3}{*}{6} & 0.580 & 0.563 & 18.1 & 8 & \multirow{3}{*}{5} & 0.194 & 0.241 \\
 & 25 & 0.616 & 0.621 & 11.4 & 8 &  & 0.632 & 0.611 & 18.1 & 6 &  & 0.398 & 0.579 \\
 & 50 & 0.770 & 0.719 & 10.5 & 6 &  & \bfseries 0.656 & 0.684 & 12.0 & 5 &  & 0.000 & 0.004 \\
\midrule
\multirow{3}{*}{R-Air} & 5 & \bfseries 0.501 & \bfseries 0.519 & 5.5 & 13 & \multirow{3}{*}{6} & 0.566 & 0.586 & 2.3 & 5 & \multirow{3}{*}{3} & 0.000 & 0.000 \\
 & 25 & \bfseries 0.534 & \bfseries 0.538 & 4.7 & 10 &  & 0.566 & 0.586 & 2.3 & 5 &  & 0.000 & 0.000 \\
 & 50 & \bfseries 0.565 & \bfseries 0.586 & 2.0 & 5 &  & 0.596 & 0.624 & 1.3 & 3 &  & 0.000 & 0.000 \\
\midrule
\multirow{3}{*}{R-Bos} & 5 & \bfseries 0.245 & 0.287 & 4.7 & 9 & \multirow{3}{*}{5} & 0.281 & 0.338 & 7.4 & 4 & \multirow{3}{*}{3} & 0.000 & 0.057 \\
 & 25 & \bfseries 0.254 & 0.290 & 3.8 & 9 &  & 0.282 & 0.338 & 7.4 & 4 &  & 0.000 & 0.021 \\
 & 50 & \bfseries 0.283 & 0.332 & 2.0 & 5 &  & 0.347 & 0.355 & 1.3 & 3 &  & 0.000 & 0.054 \\
\midrule
\multirow{3}{*}{R-Dia} & 5 & \bfseries 0.510 & 0.546 & 2.9 & 7 & \multirow{3}{*}{4} & 0.531 & 0.578 & 1.3 & 3 & \multirow{3}{*}{2} & 0.000 & 0.051 \\
 & 25 & \bfseries 0.515 & 0.546 & 2.9 & 7 &  & 0.533 & 0.577 & 1.3 & 3 &  & 0.000 & 0.046 \\
 & 50 & \bfseries 0.525 & 0.551 & 2.0 & 5 &  & 0.538 & 0.571 & 1.3 & 3 &  & 0.000 & 0.482 \\
\midrule
\multirow{3}{*}{R-Dow} & 5 & \bfseries 0.336 & \bfseries 0.357 & 3.8 & 9 & \multirow{3}{*}{4} & 0.449 & 0.445 & 2.3 & 3 & \multirow{3}{*}{2} & 0.000 & 0.000 \\
 & 25 & \bfseries 0.369 & \bfseries 0.372 & 3.8 & 9 &  & 0.449 & 0.451 & 2.3 & 3 &  & 0.000 & 0.000 \\
 & 50 & \bfseries 0.395 & \bfseries 0.418 & 2.0 & 5 &  & 0.469 & 0.466 & 1.3 & 3 &  & 0.000 & 0.000 \\
\midrule
\multirow{3}{*}{R-EnC} & 5 & \bfseries 0.099 & \bfseries 0.108 & 7.3 & 15 & \multirow{3}{*}{6} & 0.149 & 0.145 & 14.5 & 7 & \multirow{3}{*}{4} & 0.000 & 0.000 \\
 & 25 & \bfseries 0.104 & \bfseries 0.113 & 5.5 & 12 &  & 0.157 & 0.155 & 13.8 & 7 &  & 0.000 & 0.000 \\
 & 50 & \bfseries 0.117 & \bfseries 0.127 & 3.8 & 9 &  & 0.175 & 0.176 & 13.5 & 5 &  & 0.000 & 0.000 \\
\midrule
\multirow{3}{*}{R-EnH} & 5 & \bfseries 0.082 & \bfseries 0.085 & 6.0 & 13 & \multirow{3}{*}{5} & 0.130 & 0.132 & 14.5 & 8 & \multirow{3}{*}{5} & 0.000 & 0.000 \\
 & 25 & \bfseries 0.085 & \bfseries 0.087 & 4.7 & 11 &  & 0.142 & 0.141 & 13.5 & 7 &  & 0.000 & 0.000 \\
 & 50 & \bfseries 0.089 & \bfseries 0.098 & 2.9 & 7 &  & 0.164 & 0.162 & 8.4 & 5 &  & 0.000 & 0.000 \\
\midrule
\multirow{3}{*}{R-Tow} & 5 & \bfseries 0.290 & \bfseries 0.288 & 3.8 & 9 & \multirow{3}{*}{4} & 0.373 & 0.381 & 8.4 & 6 & \multirow{3}{*}{4} & 0.000 & 0.000 \\
 & 25 & \bfseries 0.298 & \bfseries 0.302 & 2.9 & 7 &  & 0.379 & 0.389 & 3.3 & 5 &  & 0.000 & 0.000 \\
 & 50 & \bfseries 0.371 & \bfseries 0.370 & 2.0 & 5 &  & 0.449 & 0.457 & 7.4 & 4 &  & 0.000 & 0.000 \\
\midrule
\multirow{3}{*}{R-Yac} & 5 & \bfseries 0.011 & \bfseries 0.014 & 11.4 & 13 & \multirow{3}{*}{9} & 0.013 & 0.017 & 7.4 & 4 & \multirow{3}{*}{2} & 0.000 & 0.000 \\
 & 25 & \bfseries 0.012 & 0.016 & 5.5 & 11 &  & 0.013 & 0.017 & 7.4 & 4 &  & 0.000 & 0.037 \\
 & 50 & 0.015 & 0.024 & 3.8 & 9 &  & 0.013 & \bfseries 0.018 & 7.4 & 4 &  & 0.006 & 0.000 \\
 \bottomrule
    \end{tabular}
\end{table}

For a same $\tau$, solutions found by $\text{NSGP}^\phi$ have typically larger $\ell$ than those found by $\text{NSGP}^\ell$.
The vice versa also holds, as can be expected. 
Notable examples appear for $\tau=25$ in S-Pa1 and R-EnC/H: $\text{NSGP}^\phi$ achieves approximately double $\ell$ compared to the $\text{NSGP}^\ell$, while the latter achieves approximately double $\phi$ compared to the former.

Regarding the training MSE, the use of $\phi$ leads to the best optimization.
This is particularly evident for $\tau=5$ where all results are significantly better when using $\text{NSGP}^\phi$, except for S-Vl4.
Using $\phi$ instead of $\ell$ has a smaller detrimental impact on finding well-fitting formulas.
A plausible explanation is that $\text{NSGP}^\phi$ explores the search space better than $\text{NSGP}^\ell$.
This hypothesis is also supported by considering the sizes of the non-dominated fronts $|\mathcal{F}|$: although the fronts are generally small for both $\phi$ and $\ell$ (reasonable because both depend on discrete properties of the solutions~\cite{smits2005pareto}), they are consistently larger when $\phi$ is used.

Less differences between $\text{NSGP}^\phi$ and $\text{NSGP}^\ell$ are significant when considering the test MSE (also due to Holm-Bonferroni correction). 
This is a normal consequence of assessing generalization as gains in training errors are lost due to (some) overfitting. 
What is important tough is that $\text{NSGP}^\phi$ remains preferable. 
For $\tau=5$ ($\tau=25$), this is the case for \num{9} (\num{7}) out of \num{12} datasets. 

\subsubsection{Qualitative results.}
We delve deeper into the results to assess the behavior of NSGP using $\phi$ and $\ell$, from a qualitative perspective. 
We consider three datasets: S-Vl4 where no version of NSGP is superior to the other; R-Bos where $\text{NSGP}^\phi$ is only better at training time; and R-EnH, where $\text{NSGP}^\phi$ is favorable also at test time. 
\Cref{fig:scatters} shows \emph{all} validation fronts obtained from the \num{50} runs, re-evaluated in terms of test MSE for both versions of NSGP, and plotted w.r.t.\ $\phi$ (left plots) and $\ell$ (right plots). 
We also show, for $\tau \in \{1, 50, 100\}$, the solutions obtained by considering always the first run (seed \num{1} in the results on our online code repository at \url{https://github.com/MaLeLabTs/GPFormulasInterpretability}).

\pgfdeclarelayer{bg}
\pgfsetlayers{bg,main}
    
\begin{figure}[t]
    \centering
    \begin{tikzpicture}[
        fnodep/.style={draw=blue,fill=white,scale=.6},
        fnodel/.style={draw=red,fill=white,scale=.6},
        flinkp/.style={draw=blue},
        flinkl/.style={draw=red}
    ]
        \pgfplotsset{
          mark layer=axis tick labels
        }
        \begin{groupplot}[
            width=0.35\linewidth,
            height=0.35\linewidth,
            every axis plot post/.append style={
                fill opacity=0.15,
                draw opacity=0.0,
                mark options={scale=0.5},
                only marks
            },
            legend image post style={
                fill opacity=1,
                draw opacity=1,
                mark options={scale=1},
            },
            group style={
                group size=2 by 3,
                horizontal sep=55mm,
                vertical sep=2mm,
                xticklabels at=edge bottom,
                yticklabels at=edge left,
            }
        ]
            \nextgroupplot[title={$\phi$},ylabel={Test MSE (S-Vl4)},legend columns=2,legend entries={NSGP$^\phi$,NSGP$^\ell$},legend style={draw=none,at={(2,1)},anchor=south}]
            \addplot[blue] table[col sep=comma,x=phi,y=mse_test]{simplified_vladislavleva4_True.txt};
            \addplot[red] table[col sep=comma,x=phi,y=mse_test]{simplified_vladislavleva4_False.txt};
            \node (vrp) at (rel axis cs:1,1) {};
            \node (vl000p) at (axis cs:6.428000,0.908000) {};
            \node (vl040p) at (axis cs:12.021000,0.712000) {};
            \node (vl100p) at (axis cs:40.065000,0.441000) {};
            \node (vp000p) at (axis cs:1.087000,0.911000) {};
            \node (vp050p) at (axis cs:1.980000,0.911000) {};
            \node (vp100p) at (axis cs:11.394000,0.595000) {};

            \nextgroupplot[title={$\ell$}]
            \addlegendimage{black,fill=gray!00,area legend}
            \addlegendimage{black,dashed,fill=gray!00,area legend}
            \addlegendimage{black,densely dotted,fill=gray!00,area legend}
            \addplot[blue] table[col sep=comma,x=n_nodes,y=mse_test]{simplified_vladislavleva4_True.txt};
            \addplot[red] table[col sep=comma,x=n_nodes,y=mse_test]{simplified_vladislavleva4_False.txt};
            \node (vrl) at (rel axis cs:0,1) {};
            \node (vl000l) at (axis cs:2.000000,0.908000) {};
            \node (vl040l) at (axis cs:5.000000,0.712000) {};
            \node (vl100l) at (axis cs:14.000000,0.441000) {};
            \node (vp000l) at (axis cs:3.000000,0.911000) {};
            \node (vp050l) at (axis cs:5.000000,0.911000) {};
            \node (vp100l) at (axis cs:8.000000,0.595000) {};
            
            \nextgroupplot[ylabel={Test MSE (R-Bos)}]
            \addplot[blue] table[col sep=comma,x=phi,y=mse_test]{simplified_boston_True.txt};
            \addplot[red] table[col sep=comma,x=phi,y=mse_test]{simplified_boston_False.txt};
            \node (brp) at (rel axis cs:1,1) {};
            \node (bl000p) at (axis cs:6.428000,0.572000) {};
            \node (bl067p) at (axis cs:1.290000,0.372000) {};
            \node (bl100p) at (axis cs:7.417000,0.214000) {};
            \node (bp000p) at (axis cs:1.087000,0.372000) {};
            \node (bp040p) at (axis cs:1.087000,0.372000) {};
            \node (bp100p) at (axis cs:15.443000,0.190000) {};

            \nextgroupplot[]
            \addplot[blue] table[col sep=comma,x=n_nodes,y=mse_test]{simplified_boston_True.txt};
            \addplot[red] table[col sep=comma,x=n_nodes,y=mse_test]{simplified_boston_False.txt};
            \node (brl) at (rel axis cs:0,1) {};
            \node (bl000l) at (axis cs:2.000000,0.572000) {};
            \node (bl067l) at (axis cs:3.000000,0.372000) {};
            \node (bl100l) at (axis cs:4.000000,0.214000) {};
            \node (bp000l) at (axis cs:3.000000,0.372000) {};
            \node (bp040l) at (axis cs:3.000000,0.372000) {};
            \node (bp100l) at (axis cs:9.000000,0.190000) {};

            \nextgroupplot[ylabel={Test MSE (R-EnH)}]
            \addplot[blue] table[col sep=comma,x=phi,y=mse_test]{simplified_energyheating_True.txt};
            \addplot[red] table[col sep=comma,x=phi,y=mse_test]{simplified_energyheating_False.txt};
            \node (erp) at (rel axis cs:1,1) {};
            \node (el000p) at (axis cs:6.428000,0.189000) {};
            \node (el057p) at (axis cs:19.671000,0.127000) {};
            \node (el100p) at (axis cs:30.857000,0.100000) {};
            \node (ep000p) at (axis cs:1.087000,0.189000) {};
            \node (ep050p) at (axis cs:2.872000,0.103000) {};
            \node (ep100p) at (axis cs:6.442000,0.094000) {};

            \nextgroupplot[]
            \addplot[blue] table[col sep=comma,x=n_nodes,y=mse_test]{simplified_energyheating_True.txt};
            \addplot[red] table[col sep=comma,x=n_nodes,y=mse_test]{simplified_energyheating_False.txt};
            \node (erl) at (rel axis cs:0,1) {};
            \node (el000l) at (axis cs:2.000000,0.189000) {};
            \node (el057l) at (axis cs:6.000000,0.127000) {};
            \node (el100l) at (axis cs:12.000000,0.100000) {};
            \node (ep000l) at (axis cs:3.000000,0.189000) {};
            \node (ep050l) at (axis cs:7.000000,0.103000) {};
            \node (ep100l) at (axis cs:15.000000,0.094000) {};
        \end{groupplot}
        
        \node(bl000f) [fnodel,anchor=east,below left=00mm and 00mm of brl] {$\sin( x_{13} )$};
        \node(bl067f) [fnodel,anchor=west,below left=08mm and 00mm of brl] {$x_6 - x_{13}$};
        \node(bl100f) [fnodel,anchor=east,below left=15mm and 00mm of brl] {$\cos( x_6 ) + x_{13}$};
        \node(bp000f) [fnodep,anchor=west,below right=00mm and 00mm of brp] {$x_{13} - x_6 $};
        \node(bp040f) [fnodep,anchor=west,below right=08mm and 00mm of brp] {$x_6 - x_{13}$};
        \node(bp100f) [fnodep,anchor=west,below right=15mm and 00mm of brp] {$\begin{aligned}( \cos( x_6 ) + x_{13} )\\- ( x_6 - \cos( x_6 ) )\end{aligned}$};
        
        \node(el000f) [fnodel,anchor=east,below left=00mm and 00mm of erl] {$\sin( x_5 )$};
        \node(el057f) [fnodel,anchor=west,below left=08mm and 00mm of erl] {$\exp( \exp( \sin( x_4 ) ) ) - x_7$};
        \node(el100f) [fnodel,anchor=east,below left=15mm and 00mm of erl] {$\begin{aligned}\exp( \exp( \sin( x_5 ) ) )\\- ( ( \sin( \cos( x_4 ) ) + x_1 )\\- x_7 )\end{aligned}$};
        \node(ep000f) [fnodep,anchor=west,below right=00mm and 00mm of erp] {$0.017 \div_p x_5$};
        \node(ep050f) [fnodep,anchor=west,below right=08mm and 00mm of erp] {$( ( -3.235 \times x_5 ) - x_7 ) - x_3$};
        \node(ep100f) [fnodep,anchor=west,below right=15mm and 00mm of erp] {$\begin{aligned}0.017 \times ( ( x_1 \times x_3 ) + ( ( x_3 + x_7 )\\+ ( 1.89 \div_p ( x_5 \div_p 1.617 ) ) ) )\end{aligned}$};
        
        \node(vl000f) [fnodel,anchor=east,below left=00mm and 00mm of vrl] {$\cos( x_3 )$};
        \node(vl040f) [fnodel,anchor=west,below left=08mm and 00mm of vrl] {$\cos( x_1 ) + \cos( x_3 )$};
        \node(vl100f) [fnodel,anchor=east,below left=15mm and 00mm of vrl] {$\begin{aligned} ( \exp( \log_p( \sin( x_5 ) ) )\\- ( \exp( \cos( x_2 ) ) + \cos( x_1 ) ) )\\- \cos( x_3 )\end{aligned}$};
        \node(vp000f) [fnodep,anchor=west,below right=00mm and 00mm of vrp] {$x_3 \times x_3$};
        \node(vp050f) [fnodep,anchor=west,below right=08mm and 00mm of vrp] {$x_3 \times ( -0.004 - x_3 )$};
        \node(vp100f) [fnodep,anchor=west,below right=15mm and 00mm of vrp] {$\log_p( ( ( x_2 \times ( x_3 \times x_4 ) ) \times x_1 ) )$};

        \begin{pgfonlayer}{bg}
            \draw[flinkl] (bl000f.west)--(bl000p.center);
            \draw[flinkl] (bl000f.east)--(bl000l.center);
            \draw[fnodel] (bl067f.west)--(bl067p.center);
            \draw[flinkl] (bl067f.east)--(bl067l.center);
            \draw[flinkl] (bl100f.west)--(bl100p.center);
            \draw[flinkl] (bl100f.east)--(bl100l.center);
            \draw[flinkp] (bp000f.west)--(bp000p.center);
            \draw[flinkp] (bp000f.east)--(bp000l.center);
            \draw[fnodep] (bp040f.west)--(bp040p.center);
            \draw[flinkp] (bp040f.east)--(bp040l.center);
            \draw[flinkp] (bp100f.west)--(bp100p.center);
            \draw[flinkp] (bp100f.east)--(bp100l.center);
            \draw[fnodel] (el000f.west)--(el000p.center);
            \draw[flinkl] (el000f.east)--(el000l.center);
            \draw[fnodel] (el057f.west)--(el057p.center);
            \draw[flinkl] (el057f.east)--(el057l.center);
            \draw[fnodel] (el100f.west)--(el100p.center);
            \draw[flinkl] (el100f.east)--(el100l.center);
            \draw[fnodep] (ep000f.west)--(ep000p.center);
            \draw[flinkp] (ep000f.east)--(ep000l.center);
            \draw[fnodep] (ep050f.west)--(ep050p.center);
            \draw[flinkp] (ep050f.east)--(ep050l.center);
            \draw[fnodep] (ep100f.west)--(ep100p.center);
            \draw[flinkp] (ep100f.east)--(ep100l.center);
            \draw[fnodel] (vl000f.west)--(vl000p.center);
            \draw[flinkl] (vl000f.east)--(vl000l.center);
            \draw[fnodel] (vl040f.west)--(vl040p.center);
            \draw[flinkl] (vl040f.east)--(vl040l.center);
            \draw[fnodel] (vl100f.west)--(vl100p.center);
            \draw[flinkl] (vl100f.east)--(vl100l.center);
            \draw[fnodep] (vp000f.west)--(vp000p.center);
            \draw[flinkp] (vp000f.east)--(vp000l.center);
            \draw[fnodep] (vp050f.west)--(vp050p.center);
            \draw[flinkp] (vp050f.east)--(vp050l.center);
            \draw[fnodep] (vp100f.west)--(vp100p.center);
            \draw[flinkp] (vp100f.east)--(vp100l.center);
        \end{pgfonlayer}

    \end{tikzpicture}
    \caption{
        Scatter plots of validation fronts as re-evaluated on the test set for all \num{50} runs, in terms of $\phi$ (left column) and $\ell$ (right column).
        Formulas in the middle are picked from the front of run \num{1}, using $\tau = 1$ (bottom), $50$ (middle), $100$ (top). Note that $x_{13}-x_{6}$ and $x_{6}-x_{13}$ (R-Bos) are equivalent due to linear scaling.
    }
    \label{fig:scatters}
\end{figure}
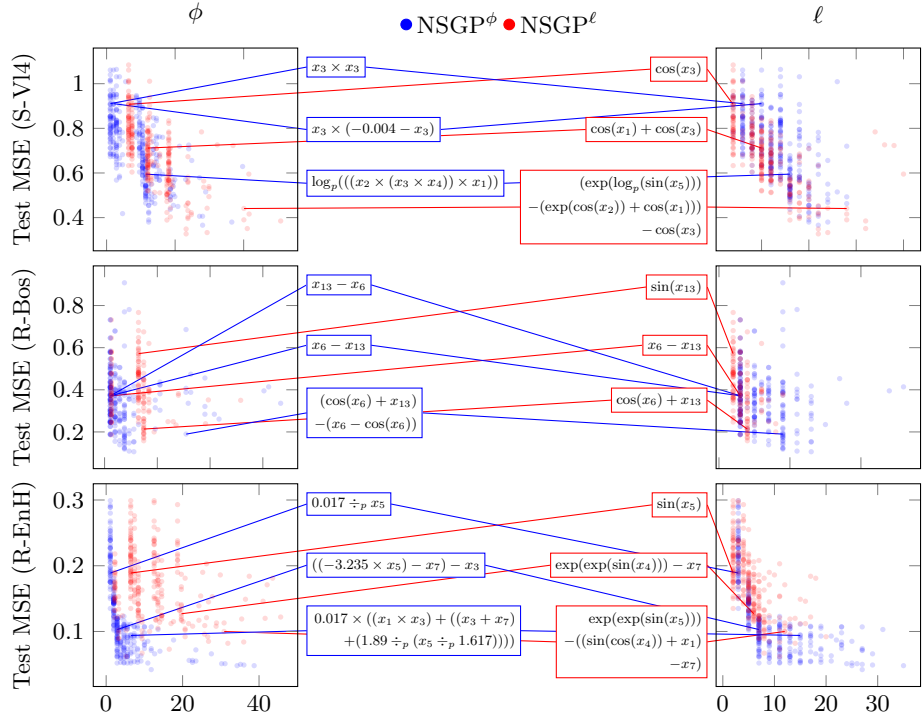

The scatter plots show that, in general, $\text{NSGP}^\phi$ obtains more points with small test MSE.
This is most evident for R-EnH, where the results are found to be statistically significant. 
Note how, instead, this is not the case for $\tau=100$ in S-Vl4, where in fact the use of $\phi$ is no better than the use of $\ell$ (see \Cref{tab:percentiles-scores}).

By visually inspecting the formulas, we find results that are in line with what found in \Cref{tab:percentiles-scores}. 
Formulas found by $\text{NSGP}^\phi$ with small MSE ($\tau=1,50$) can often be (slightly) longer than their counterpart found by $\text{NSGP}^\ell$ (except for S-Vl4), however, they typically contain less non-arithmetic operations, and less of their compositions.
Even for very small formulas, those found $\text{NSGP}^\ell$ rely more on non-arithmetic operations, meaning these operations help achieving small MSE, at least up to the validation stage. 
All in all, the most complex formulas found by $\text{NSGP}^\phi$ are either more easily or similarly interpretable than those found by $\text{NSGP}^\ell$.

\section{Discussion}
To realize our data-driven approach, we relied on a survey aimed at measuring human-interpretability of mathematical formulas.
While we did our best to design a survey that could gather useful human feedback, a clear limitation of our work is the relatively small number of survey respondents (\num{334}), which in turn led to obtaining a relatively small dataset (\num{73} samples, \num{4} formula features).
Fitting of a high-bias (linear) model resulted in a decent test $R^2$ of $0.5$, while having the model be interpretable itself.
Still, the model need not be interpretable.
With more data available in the future, we will investigate the use of a larger number of more sophisticated features~\cite{maruyama2012cortical}, and the use of more complex (possibly even black-box) models. 
Moreover, our approach can also be investigated for ML models other than formulas (e.g., decision trees).

In terms of results with NSGP, we found that $\phi$ allows the discovery of good solutions w.r.t.\ the competing objective, i.e., the MSE, better than $\ell$. 
We also found that using $\phi$ leads to the discovery of larger fronts. 
There is no reason to expect this outcome beforehand, as $\phi$ was not designed to achieve this.
We believe these findings boil down to one fundamental reason: diversity preservation.
Because the estimation of $\phi$ relies on more features compared to the measurement of $\ell$, more solutions can co-exist that do not dominate each other.
Hence, the use of $\phi$ fares better against premature convergence~\cite{squillero2016divergence}.

Regarding the examples of formulas we obtained, one may think that $\phi$ leads to arguably more interpretable formulas than $\ell$ simply because it accounts for non-arithmetic operations (and their composition). 
In fact, we agree that $\ell$ is simplistic.
However, we believe that minimizing $\ell$ represents one of the first baselines to compare against (and it was the only one we found being used to specifically promote interpretability~\cite{lensen2020genetic}), and that designing a competitive baseline is non-trivial.
We will investigate this further in future work.

What about formula simplification? 
We did not present results regarding formulas after a simplification step. 
We attempted to use the sympy library~\cite{sympy} to assess the effect of formula simplification during the evolution, but to no avail as runtimes exploded.
Moreover, we looked at what happens if we simplify (with sympy) the formulas in the final front, and re-compute their $\phi$ and $\ell$.
Results were mixed.
For example, regarding the three datasets of \Cref{fig:scatters}, re-measuring $\phi$ and $\ell$ after simplification led to a mean improvement ratio of \num{1.08} (\num{1.17}) and \num{1.00} (\num{1.00}) respectively, when all (only the most complex) formulas from the fronts were considered. 
Hence, the use of $\phi$ seems more promising than $\ell$ w.r.t.\ simplification. 
Yet, as improvements were small (also in visual assessments), further investigation will be needed.

\section{Conclusion}
We presented a data-driven approach to learn, from responses to a survey on mathematical formulas we designed, a model of interpretability. 
This model is itself an interpretable (linear) formula, with reasonable properties. 
We plugged-in this model within multi-objective genetic programming to promote formula interpretability in symbolic regression, and obtained significantly better results when comparing with traditional formula size minimization. 
As such, our approach represents an important step towards better interpretable machine learning, especially by means of multi-objective evolution.
Furthermore, the model we found can be used as a proxy of formula interpretability in future studies.

\section*{Acknowledgments and author contributions}
We thank the Maurits en Anna de Kock Foundation for financing a high-performance computing system that was used in this work.
Author contributions, in order of importance, follow. Conceptualization: M.V.; methodology: M.V., E.M.; software: M.V., A.D.L., F.R.; writing: M.V., E.M., A.D.L., F.R.

\bibliographystyle{splncs04}
\bibliography{bibliography}

\end{document}